\documentclass[runningheads]{llncs}
\pdfoutput=1
\usepackage{graphicx}
\usepackage{xcolor}
\usepackage{amsmath}
\usepackage{amssymb}
\usepackage{booktabs}
\usepackage{comment}
\usepackage{lipsum}
\usepackage{bbm}
\usepackage[misc]{ifsym}
\usepackage{pdfpages}

\usepackage{hyperref}
\usepackage{ifthen}

\usepackage{xspace}
\newcommand{\todocite}[1]{\textcolor{red}{[cite!\ifthenelse{\equal{#1}{}}{}{ #1}]}\xspace}
\newcommand{\todo}[1]{\textcolor{red}{(todo\emph{\ifthenelse{\equal{#1}{}}{}{ #1}!)}}\xspace}

\DeclareMathOperator*{\argmin}{arg\,min}

\usepackage{color}

\definecolor{ben}{rgb}{0.9,0.,0.5}

\definecolor{vincent}{rgb}{0.,0.5,0.5}

\definecolor{alex}{rgb}{0.,0.,0.8}

\definecolor{hayoung}{rgb}{0.5,0.2,0.8}

\definecolor{mahdi}{rgb}{0.1,0.7,0.9}

\usepackage{siunitx}
\DeclareSIUnit{\nothing}{\relax}
\usepackage{multirow}

\makeatletter
\newcommand{\printfnsymbol}[1]{%
  \textsuperscript{\@fnsymbol{#1}}%
}
\makeatother

\begin{document}
\title{S3M: Scalable Statistical Shape Modeling through Unsupervised Correspondences}
\titlerunning{S3M: Scalable Statistical Shape Modeling}

\author{
Lennart Bastian\thanks{denotes equal contribution. \qquad \qquad \Letter: \email{lennart.bastian@tum.de}} \textsuperscript{\Letter} \and 
Alexander Baumann\printfnsymbol{1} \and
Emily Hoppe \and
Vincent Bürgin \and \\
Ha Young Kim \and
Mahdi Saleh \and
Benjamin Busam \and
Nassir Navab
}

\authorrunning{L. Bastian et al.}
\institute{Computer Aided Medical Procedures, Technical University Munich, Germany}
\maketitle              %
\begin{abstract}
Statistical shape models (SSMs) are an established way to represent the anatomy of a population with various clinically relevant applications.
However, they typically require domain expertise, and labor-intensive landmark annotations to construct.
We address these shortcomings by proposing an unsupervised method that leverages deep geometric features and functional correspondences to simultaneously learn local and global shape structures across population anatomies.
Our pipeline significantly improves unsupervised correspondence estimation for SSMs compared to baseline methods, even on highly irregular surface topologies.
We demonstrate this for two different anatomical structures: the thyroid and a multi-chamber heart dataset. 
Furthermore, our method is robust enough to learn from noisy neural network predictions, potentially enabling scaling SSMs to larger patient populations without manual segmentation annotation. The code is publically available at:\\ \href{https://github.com/alexanderbaumann99/S3M
}{https://github.com/alexanderbaumann99/S3M}

\end{abstract}
\section{Introduction}

Statistical shape models (SSMs) are a powerful tool to characterize anatomical variations across a population. 
They have been widely used in medical image analysis and computational anatomy to represent organ structures, with numerous clinically relevant applications such as clustering, classification, and shape regression \cite{heimannStatisticalShapeModels2009,bhalodia2021deepssm,adams2022images}.
SSMs are generally represented by point-wise correspondences between shapes \cite{cootesTrainingModelsShape1992,agrawal2021learning}, or deformation fields to a pre-defined template \cite{cootes2008diffeomorphic,ludke2022landmark}.
Despite the existence of implicit models \cite{adams2022images}, abstracting shape correspondences in the form of linear point distribution models (PDM) constitutes an appealing and interpretable way to represent shape distributions \cite{adams2023spatiotemporal}.
Furthermore, many implicit models still rely on correspondence annotations during training \cite{bhalodia2021deepssm,adams2022images}.

Creating SSMs is cumbersome and intricate, as significant manual human annotation is necessary.
Domain experts typically first segment images in 3D.
The labeled 3D organ surfaces must then be aligned and brought into correspondence, typically achieved through deformable image registration methods using manual landmark annotations \cite{banerjee2022automated}.
This is labor-intensive and error-prone, potentially inducing bias into downstream SSMs and applications \cite{zhang2020disentangling}.

Unsupervised methods have been proposed to estimate correspondences for SSMs \cite{catesShapeWorks2017,klatzowMMatch3DShape2022,agrawal2021learning}.
However, they typically require precisely segmented and smooth surfaces to generate accurate inter-organ correspondences.
ShapeWorks has been established to produce high-quality correspondences on several organs such as femurs or left atria \cite{adams2020uncertain,bhalodia2021deepssm,adams2023spatiotemporal}.
However, as domain experts carefully curate most medical datasets, the robustness of such methods has not yet been thoroughly evaluated concerning label noise and segmentation inaccuracies.
The main obstacles that prevent scaling SSMs to larger patient populations are unsupervised correspondence methods that can handle topological variations in noisy annotations, such as those produced by inexperienced annotators or predictions from deep neural networks.
Robust methods to deal with these obstacles are required.

To address these challenges, we propose S3M, which leverages unsupervised deep geometric features while incorporating a global shape structure.
Geometric Deep Learning (GDL) provides techniques to process 3D shapes and geometries, which are robust to noise, 3D rotations, and global deformations. 
We utilize graph neural networks (GNN) and functional mappings to establish dense surface correspondences of samples without supervision.
This approach has significant clinical implications as it enables automatically representing anatomical shapes across large patient populations without requiring manual expert landmark annotations.
We demonstrate that our proposed method creates objectively superior SSMs from shapes with noisy surface topologies.
Moreover, it accurately corresponds regions of complex anatomies with mesh bifurcations such as the heart, which could ease the modeling of inter-organ relations~\cite{cerrolaza2019computational}.

Our contributions can be summarized as follows:
\begin{itemize}
    \item We propose a novel unsupervised correspondence method for SSM curation based on geometric deep learning and functional correspondence.
    \item S3M exhibits superior performance on two challenging anatomies: thyroid and heart. It generates objectively more suitable SSMs from noisy thyroid labels and challenging multi-chamber heart reconstructions.
    \item To pave the way for unsupervised SSM generation in other medical domains, we open-source the code of our pipeline.
\end{itemize}
\section{Related Work}

\begin{figure}[t]
    \centering
    \includegraphics[width=\textwidth]{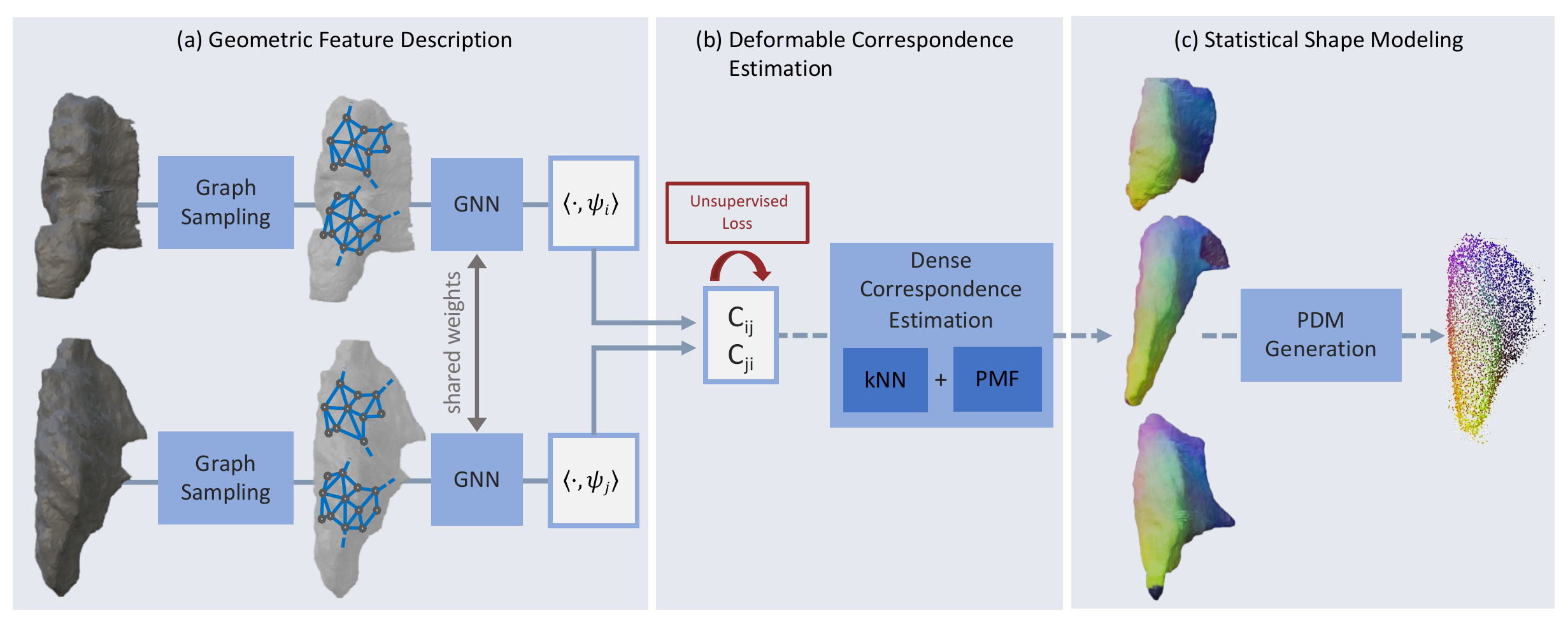}
    \caption{Our proposed method for unsupervised SSM curation. 
    (a) We use a Siamese GNN as shape descriptor and project the extracted features onto the LBO eigenfunctions $\psi$ to obtain spectral representations.
    (b) From these, we optimize a functional mapping between pairs of shapes with an unsupervised loss. 
    Gradients are backpropagated to the geometric descriptors.
    (c) During inference, the dense correspondences are estimated 
    between pairs of shapes 
    based on the learned population parameters, which are then used to construct an SSM.}
    \label{fig:overview}
\end{figure}

\noindent\textbf{Point Distribution Models.} 
A population of shapes must be brought into correspondence to construct a PDM. 
This has been traditionally achieved through pair-wise registration methods \cite{heimannStatisticalShapeModels2009,goparaju2022benchmarking}.
However, pairwise approaches can admit bias as they neglect the population during correspondence generation \cite{goparaju2022benchmarking}.
More recently, group-wise optimization methods such as ShapeWorks \cite{catesShapeWorks2017} have been adopted as they jointly optimize over a cohort, overcoming such biases. 
They demonstrate superior prediction of clinically relevant anatomical variations \cite{goparaju2022benchmarking}. 
Furthermore, generic models that can perform well across various organs are sought after.

\noindent\textbf{Graph Neural Networks} (GNNs) have been used to enable structural feature extraction through message passing.
They constitute a powerful tool to process 3D data and extract geometric features~\cite{wang2019dynamic,saleh2020graphite,saleh2022bending} which can be useful for disease prediction~\cite{kazi2019inceptiongcn,parisot2017spectral}.
Other medical applications involve brain cortex mesh reconstruction~\cite{bongratz2022vox2cortex} and 3D organ-at-risk (ORA) segmentation~\cite{henderson2022automatic}.
We use GNNs for deformable 3D organ shapes and learn to estimate dense correspondences in the presence of noise and anatomical variations.

\noindent\textbf{Functional Correspondence.} Functional maps abstract the notion of point-to-point correspondences by corresponding arbitrary functions, such as texture or curvature, across shapes.
They are extensively used to estimate dense correspondences across deformable shapes \cite{ovsjanikovFunctionalMapsFlexible} and can be incorporated into learning frameworks \cite{litany2017deep,donati2020deep}.
These methods are typically evaluated on synthetic meshes with dense annotations and limited variable surface topology. 
In contrast, medical shapes exhibit higher variability, requiring robust surface representation for reliable correspondence matching.
More recently, unsupervised functional correspondence models have been proposed \cite{roufosseUnsupervisedDeepLearning2019,cao2022unsupervised}.
These methods demonstrate strong performance on synthetic data without manual correspondence annotations.
They extract features from the surface geometry using hand-crafted descriptors such as SHOT \cite{hutchisonUniqueSignaturesHistograms2010}, wave-kernel signatures (WKS) \cite{aubryWaveKernelSignature2011} or heat-kernel signatures (HKS) \cite{bronstein2010scale}. 
The extracted features are then typically refined and projected onto the Laplace-Beltrami Operator (LBO) eigenfunctions \cite{ovsjanikovFunctionalMapsFlexible}.
$\mu$Match \cite{klatzowMMatch3DShape2022} recently leveraged such an unsupervised approach in the medical domain.
They employ handcrafted features to extract representations from shapes with a relatively smooth surface topology; however, they fail for shapes with high degrees of surface noise or label inconsistencies.
To scale SSM curation to larger datasets encompassing population variance, our method must be robust to a more variable and complex surface topology.

\section{Method}

In the following, we propose a method to establish an SSM as a Point Distribution Model (PDM), illustrated in \autoref{fig:overview}.
Robust local features from the surface mesh are extracted using GNNs.
These features are then projected onto the truncated eigenspace of the Laplace-Beltrami Operator using $m=20$ eigenfunctions \cite{ovsjanikovFunctionalMapsFlexible}.
We perform post-processing with a Product Manifold Filter (PMF) \cite{vestnerProductManifoldFilter2017} to obtain bijective correspondences for SSM generation.
The shape model is subsequently created by aggregating correspondences across a dataset of predicted correspondences using the eigendecomposition of the covariance matrix.

\noindent\textbf{Geometric Feature Description.}
\label{sec:geometric_features}
Handcrafted descriptors \cite{hutchisonUniqueSignaturesHistograms2010,aubryWaveKernelSignature2011,bronstein2010scale} are unable to represent the complex surface topology of medical data adequately.
To cope with surface artifacts and irregular morphologies, we use a graph-based descriptor~\cite{saleh2020graphite}.
We first extract a surface mesh from a 3D volumetric grid using marching cubes~\cite{lorensen1987marching}. 
Graph nodes are defined as the mesh's vertices; edges are obtained using a k-nearest neighbor search with $k=10$. 
Node features are given by spatial xyz-coordinates.
The graph is fed into three topology adaptive layers~\cite{du2017topology} using graph convolutions with a specific number of hops to define the number of nodes a message is passed to.
Increasing the number of hops (we use 1, 2, 3 hops per layer, respectively) increases the receptive field, incorporating features from more distant nodes.
Finally, features pass a linear layer before being projected onto the Laplace-Beltrami eigenfunctions.

\noindent\textbf{Deformable Correspondence Estimation.}
\label{sec:defo_correspondence}
PDMs require correspondences between samples to model the statistical distribution of the organ.
Inspired by methods for geometric shape correspondence, we propose to estimate a functional mapping $T$ to correspond high-level semantics from two input shapes, $X_i$ and $X_j$.
The LBO extends the Laplace operator to Riemannian manifolds, capturing intrinsic characteristics of the shape independent of its position and orientation in Euclidean space.
It can be efficiently computed on a surface mesh using, for example, the cotangent weight discretization scheme \cite{crane2018discrete}.
This results in a matrix representation of the LBO from which one can then calculate the associated eigenfunctions $\psi_i\in\mathbf{R}^{n\times m}$ for a shape $X_i \in \mathbf{R}^{n\times 3}$.
Given a feature vector $D_i$ extracted from a surface mesh of shape $X_i$ and a neural network $T_\phi$, we can approximate a functional mapping between shapes by solving the following optimization problem: 
\begin{align}
    \label{eq:loss}
    &\min_{\phi}\sum_{(X_i,X_j)} \mathcal{L}(C_{ij},C_{ji})\text{\quad where \quad } C_{ij} = \argmin_{C}\lVert CA_{T_\phi(D_i)}-A_{T_\phi(D_j)} \rVert
\end{align}
Here, $A_{T_\phi(D_i)}\in\mathbf{R}^{m\times m}$ denotes the transformed descriptor, written in the basis of the LBO eigenfunctions of shape $X_i$ and $C_{ij}\in\mathbf{R}^{m\times m}$ represents the optimal functional mapping from the descriptor space of $X_i$ to the one of $X_j$.
Inspired by existing works on shape correspondence \cite{cao2022unsupervised,roufosseUnsupervisedDeepLearning2019,sharma2020weakly}, our loss function enforces four separate characteristics on the learned functional mapping, including bijectivity, orthogonality, and isometric properties. We refer to the supplementary materials for the complete definition.
Notably, none of these losses uses ground truth correspondences, making the entire process unsupervised.

\begin{figure}[!t]
    \centering
    \includegraphics[width=\textwidth]{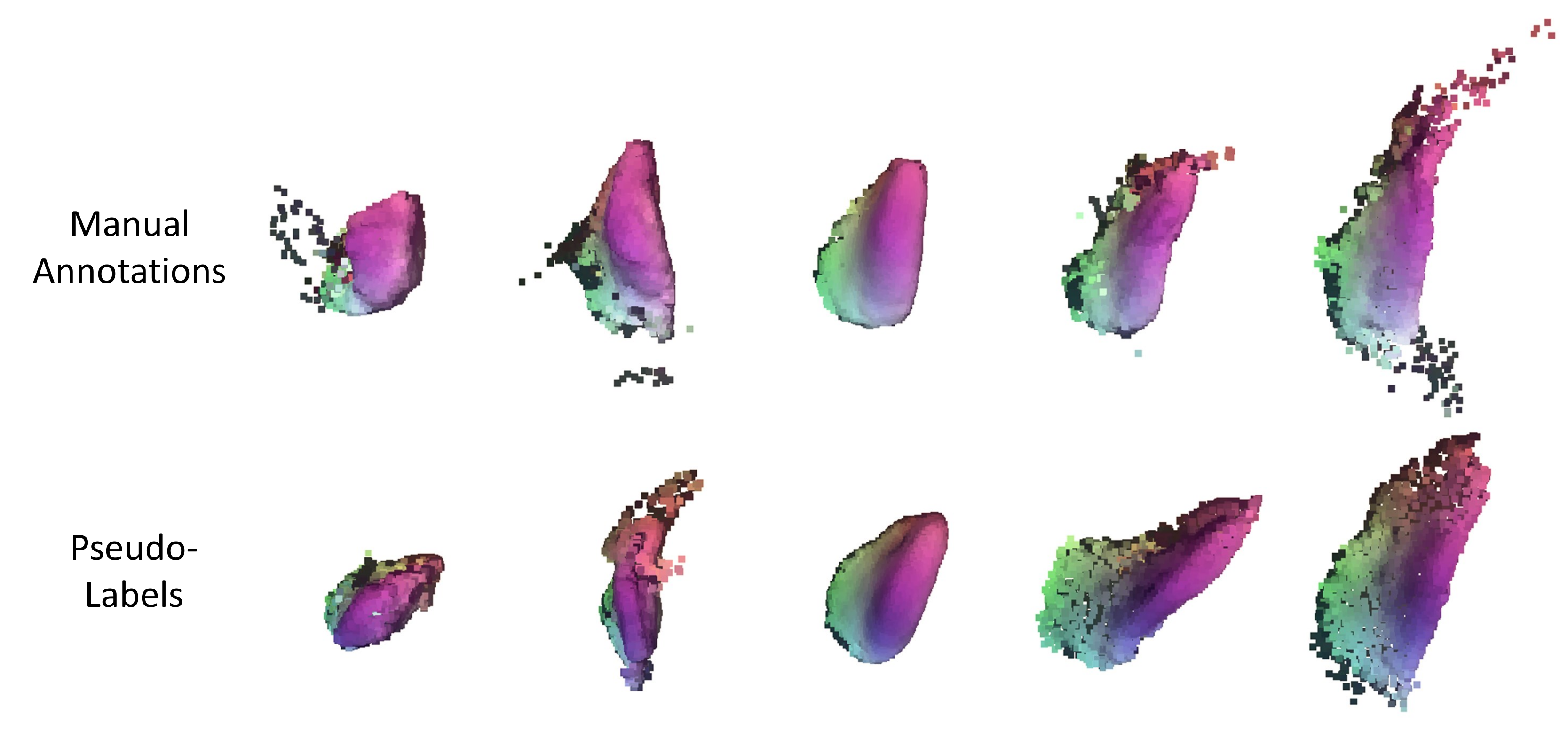}
    \caption{PDM results from the proposed method S3M on the thyroid dataset \cite{kronkeTracked3DUltrasound2022}. The top row depicts the PDM generated from manual annotations and the bottom row from network pseudo-labels.
    From left to right, we depict -3$\sqrt{\lambda_1}$, -3$\sqrt{\lambda_2}$, the mean shape, +3$\sqrt{\lambda_2}$, +3$\sqrt{\lambda_1}$, with $\lambda_i$ the eigenmode corresponding to the $i$-th largest eigenvalue.
    Similar colors indicate corresponding regions predicted by the model.}
    \label{fig:correspondance}
\end{figure}

\noindent\textbf{Training and Inference.}
During training, two shapes are sampled from the dataset, and the pipeline is optimized with \autoref{eq:loss}.
We increase model robustness by augmenting with rotations and small surface deformations.
The point cloud is sub-sampled in each training iteration using farthest point sampling with random initialization. During inference, our model predicts pairwise correspondences.
To accumulate these over an entire dataset of $N$ shapes, we choose a template shape $X_T = \argmin_{X_i} \sum_{j=0}^{N} \mathbbm{1}_{i \neq j} \mathcal{L}\left(C_{ij},C_{ji}\right)$ as the instance with the lowest average loss to all other shapes in the dataset. 
As in \cite{ovsjanikovFunctionalMapsFlexible}, we extract point-to-point correspondences between the template $X_T$ and another shape $X_i$ by matching the transformed LBO eigenfunctions of $X_i$, namely $C_{iT}\psi_i$, with the LBO eigenfunctions of the template $\psi_T$ using nearest neighbors. 
As PDMs require bijective correspondences, we subsequently post-process the results with PMF \cite{vestnerProductManifoldFilter2017}.

\noindent\textbf{Statistical Shape Modeling.}
\label{sec:ssm_method}
We use the PDM \cite{cootesTrainingModelsShape1992} as the underlying method of the shape model. It takes input points of the form $X\in\mathbf{R}^{N\times d}$, where $N$, $d$ are the number of shape samples and coordinates per shape, respectively. It returns a multi-variate normal distribution. In our case, $d=3 n$ given each shape has $n$ points.
We calculate the mean shape $\bar{X}$ and the empirical covariance matrix $S = \text{cov(X)}$ over the $N$ samples \cite{cootesTrainingModelsShape1992}.
Since $S$ has rank $N-1$, it has $N-1$ real eigenvectors $v_j$ with eigenvalues $\lambda_j$. If we consider the sum
\begin{align}
    s = \bar{X} + \sum_{j=1}^{N-1}{\alpha_j \lambda_j v_j},\qquad \alpha_j\sim\mathcal{N}(0,1)
    \label{ssmSampleStdNormal}
\end{align}
then 
$s\sim\mathcal{N}\left(\bar{X},S\right)$, which is
the desired distribution of the model. 
For the above, the points must be in correspondence across the samples. 
We thus use the correspondences generated in section \ref{sec:defo_correspondence} to construct the PDM.
\section{Experiments}
\label{heart_data}
All experiments are carried out using two publicly available datasets: thyroid ultrasound scans and heart MRI acquisitions.
Our model is implemented in PyTorch 1.12 using CUDA 11.6. Training takes between $2.5 - 3$ hours on an Nvidia A40 GPU and inference about $0.71$ seconds for a pair of shapes.
We use publically available implementations for all baseline methods.

\begin{figure}[t]
    \centering
    \includegraphics[width=\textwidth]{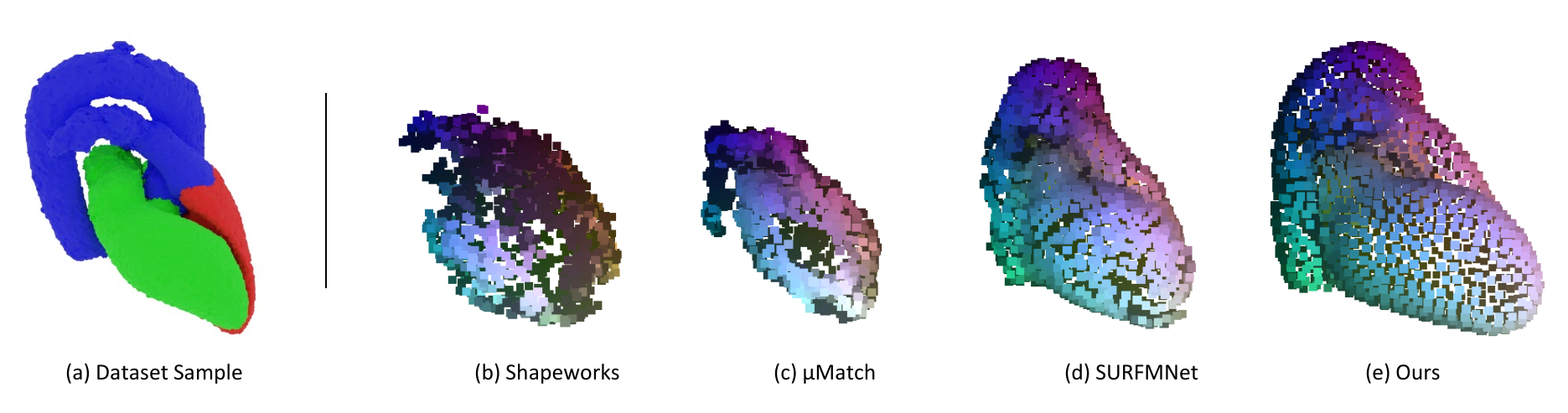}
    \caption{\textbf{Qualitative Analysis of Whole-Heart SSMs.} A sample from the heart dataset (a).
    Composition 1 (right ventricle) is denoted in red. Composition 2 (both ventricles and atria) combines red and green regions. Composition 3 additionally incorporates the vessels, denoted in blue
    The predicted SSM mean shapes for composition 3 are portrayed for ShapeWorks (b), $\mu$Match (c), SURFMNet (d), and S3M (e).}
    \label{fig:heart}
\end{figure}

\noindent\textbf{Thyroid Dataset (SegThy) \cite{kronkeTracked3DUltrasound2022}.}
The dataset comprises 3D freehand US scans of healthy thyroids from 32 volunteers aged 24-39.
For each volunteer, three physicians acquired three scans each.
Ultrasound scans generally exhibit noise induced by physical properties such as phase aberrations and attenuation. 
This leads to label inconsistencies or topological irregularities that pose a challenge for shape modeling (see \autoref{fig:correspondance}).
US sweeps were compounded to a 3D grid of resolution $0.12 \times 0.12 \times 0.12 \text{mm}^3$.
A single scan from each of the 16 volunteers was manually annotated by experts (ground truth) and used to train QuickNAT \cite{kronkeTracked3DUltrasound2022}.
The remaining scans were pseudo-labeled through QuickNAT segmentation predictions exhibiting moderate degrees of noise and inaccuracies (dice score of 0.94 \cite{kronkeTracked3DUltrasound2022}).
We divide the dataset into manual and pseudo-label predictions and evaluate them separately.
The pseudo-label experiment evaluates the model's performance under topological noise and inaccuracies and is limited to 100 scans due to ShapeWorks memory constraints \cite{gutierrez2019learning}.
We extract a surface mesh for each scan using marching cubes and subsample the meshes to 5000 vertices.

\noindent\textbf{Heart Dataset \cite{heartdata}.} 
The data constitutes 30 MRI scans from a single cardiac phase of the heart. 
Each image has a voxel resolution of $1.25 \times 1.25 \times 2.7 \text{mm}^3$.
Segmentation is carried out using an automated method, with subsequent manual corrections by domain experts. 
Labels are provided for: the right/left ventricle, right/left atrium, aorta, and pulmonary artery.
We evaluate the capability of the models to reconstruct complex organs using three hierarchical compositions of the heart chambers.
\textit{Composition 1} consists of the right ventricle, 
\textit{Composition 2} of the left and right atrium and ventricle, and \textit{Composition 3} of the whole heart, including the aorta and pulmonary artery.

\begin{table}[h]
    \centering
    \caption{\textbf{SSM quality metrics for the Thyroid dataset \cite{kronkeTracked3DUltrasound2022}}}
    \label{tab:thy_res}
    \resizebox{0.99\textwidth}{!}{
    \setlength{\tabcolsep}{0.8em} %
    {\renewcommand{\arraystretch}{1.5} %
\begin{tabular}{l|cc|cc}
  \toprule
  {Metrics} & \multicolumn{2}{c|}{Ground-Truth Segmentation} & \multicolumn{2}{c}{Network Pseudo-Labels} \\
  & {Generality [mm] $\downarrow$} & {Specificity [mm] $\downarrow$} & {Generality [mm] $\downarrow$} & {Specificity [mm] $\downarrow$} \\
  \midrule
  SURFMNet & 2.20 $\pm$ 0.20 & 3.20 $\pm$ 0.29 & - & - \\
  $\mu$Match & 1.92 $\pm$ 0.07 & 2.81 $\pm$ 0.18 & 1.90 $\pm$ 0.08  & 2.84 $\pm$ 0.10 \\
  Shapeworks & 1.94 $\pm$ 0.27 & 1.81 $\pm$ 0.06 & 1.60 $\pm$ 0.04 & 1.75 $\pm$ 0.07 \\
  S3M (Ours) & \textbf{1.25 $\pm$ 0.11} & \textbf{1.59 $\pm$ 0.06} & \textbf{0.95 $\pm$ 0.07} & 1.84 $\pm$ 0.08\\
  \bottomrule
\end{tabular}
    }
    }
\end{table}

\noindent\textbf{SSM Evaluation.}
We compare Shapeworks \cite{catesShapeWorks2017}, $\mu$Match~\cite{klatzowMMatch3DShape2022} and SURFMNet \cite{roufosseUnsupervisedDeepLearning2019}, with S3M.
A four-fold cross-validation is employed. 
SURFMNet, $\mu$Match, and S3M are trained on the training folds and correspondences are predicted on the training and validation set. 
Since the particle-based optimization of Shapeworks does not generalize to unseen data, it uses all folds for correspondence estimation.
The SSM is built using correspondences from the training set, and evaluated with respect to two standardized metrics: generality and specificity \cite{daviesLearningShapeOptimal}. 
For generality, we measure how well the SSM can represent unseen instances from the fourth fold through the Chamfer distance between the original shape and its SSM reconstruction.
Specificity indicates how well random samples from the SSM represent the training data. 
We sample from the PDM 1000 times and calculate each sample's minimum Chamfer distance to the training shapes.
Generalization and specificity are reported in mm.
Numbers in bold indicate statistically significant results by a one-sided t-test ($p < 0.05$).

\section{Results \& Discussion}

\noindent\textbf{Experiment 1: Thyroid SSM.}
Table \ref{tab:thy_res} depicts the performance for all methods on the thyroid shapes. 
SURFMNet results for the thyroid pseudo-labels are omitted, as the method did not converge.
Consistent trends can be observed across all methods for both sets of thyroid labels.
The two existing functional map-based methods were outperformed by Shapeworks, while the proposed S3M exceeded the latter's scores.
The descriptor is the most significant difference between the three learned functional map methods.
Hand-crafted shape descriptors like SHOT and a simple fully-connected residual network architecture do not adequately represent thyroids' noisy and heterogeneous shapes. 

Our proposed method significantly outperformed Shapeworks in all metrics except the specificity of pseudo-labeled thyroids, where the results are not statistically significant. 
This was despite the advantage of optimizing correspondences across all shapes in training and validation.
S3M can better cope with topological noise and generalizes to unseen samples, demonstrating potential in scaling SSM generation to larger datasets.
Furthermore, it does not suffer from increasing memory requirements with the number of samples.

\begin{table}[h]
    \centering
    \caption{\textbf{Whole Heart Statistical Shape Modeling}}
    \label{tab:heart_res}
    \resizebox{0.99\textwidth}{!}{
    \setlength{\tabcolsep}{0.5em} %
    {\renewcommand{\arraystretch}{1.5} %
    \begin{tabular}{l | cc | cc | cc}
       \toprule
      {Metrics} & \multicolumn{2}{c|}{Composition 1} & \multicolumn{2}{c|}{Composition 2} & \multicolumn{2}{c}{Composition 3} \\
      & {Generality [mm] $\downarrow$} & {Specificity [mm] $\downarrow$} & {Generality [mm] $\downarrow$} & {Specificity [mm] $\downarrow$} & {Generality [mm] $\downarrow$} & {Specificity [mm] $\downarrow$}  \\
      \midrule
      SURFMNet  & $1.10\pm 0.39$ & $1.79\pm 0.87$ & $1.37 \pm 0.16$ & $2.00 \pm 0.30$ & $1.71\pm 0.17 $ & $2.54 \pm 0.36 $ \\     
      $\mu$Match  & $3.39 \pm 0.26 $ & $4.65 \pm 0.16 $ & $2.82 \pm 0.14$ & $4.81 \pm 0.41$ & $3.30\pm 0.07$ & $5.80\pm 0.08$ \\
      ShapeWorks & $0.89\pm 0.08$ & $1.40\pm 0.04$ & $2.57 \pm 0.17$ & $3.60 \pm 0.06$ & $3.07\pm 0.19$ & $4.99\pm 0.22$ 
      \\
      S3M (Ours) & $0.85\pm 0.07 $ & $1.30\pm 0.01$ & $1.30 \pm 0.13$ & $1.72 \pm 0.05$ & $1.63 \pm 0.17$ & $2.14\pm 0.07$ \\
      \bottomrule
    \end{tabular}
    }
    }
\end{table}

\noindent\textbf{Experiment 2: Whole Heart SSM.}
\autoref{tab:heart_res} depicts the results of the different models on the three heart chamber compositions as previously defined.
For the single-organ right atrium (composition 1), our proposed method fares comparably to ShapeWorks.
For the more complex compositions 2 and 3, we observe larger increases in generalization and specificity for Shapeworks.
$\mu$Match fails to create a convincing SSM for any heart composition.
Interestingly, SURFMNet can represent the more complex compositions 2 and 3 better than ShapeWorks, showing the strength of functional maps at representing complex high-level structures.
S3M still exceeded the performance of SURFMNet, possibly due to the graph descriptor being better able to represent the surface topology.

From the qualitative results in \autoref{fig:heart}, it becomes more apparent that ShapeWorks does not generate an adequate SSM for the more complex compositions.
This further supports our proposed method's ability to learn correspondences for intricate and complex surface topologies, even consisting of meshes with bifurcations.
The flexibility of our surface representation enables unsupervised correspondence estimation from multiple hierarchical sub-shapes, which is invaluable in multi-organ modeling such as for the heart \cite{banerjee2022automated,cerrolaza2019computational}.

\noindent\textbf{Experiment 3: Thyroid Pseudo-label Generalization}

To further highlight the proposed methods' robustness to network-generated segmentation labels, we additionally measure the reconstruction ability of SSMs created from pseudo-labels on manually annotated thyroid labels under the Chamfer distance (Ours: $1.05\pm 0.10$mm, Shapeworks: $1.84\pm 0.40$mm). 
Notably, the proposed PDM on pseudo-labels generalizes better than the SSM built on few manual labels ($1.25\pm 0.11 $mm; see Table \ref{tab:thy_res}), suggesting that more data can improve the SSM even if the labels are inaccurate. 
This is further supported by differences in shape (suppl. fig. 1); the SSM's mean shape generated from pseudo-labels approximates the mean shapes on GT labels (and thus, the true organ shape) more closely.
\section{Conclusion}
We present an unsupervised approach for learning correspondences between shapes that exhibit noisy and irregular surface topologies. 
Our method leverages the strengths of geometric feature extractors to learn the intricacies of organ surfaces, as well as high-level functional bases of the Laplace-Beltrami operator to capture more extensive organ semantics.
S3M outperforms existing methods on both manual labels, and label predictions from a network, demonstrating the potential to scale existing SSM pipelines to datasets that encompass more substantial population variance without additional annotation burden.
Finally, we show that our model has the potential to learn correspondences between complex multi-organ shape hierarchies such as chambers of the heart, which would ease the manual burden of SSM curation for structures that currently still require meticulous manual landmark annotations.\\

\noindent\textbf{Acknowledgements} This work originated as a TUM Data Innovation Lab project through the Munich Data Science Institute. Vincent Bürgin is supported by the DAAD program Konrad Zuse Schools of Excellence in Artificial Intelligence, sponsored by the Federal Ministry of Education and Research. The authors declare no additional conflicts of interest.

\bibliographystyle{splncs04}
\bibliography{references}

\end{document}


\title{S3M: Supplementary Materials}
\begin{center}
  \Large\bfseries\boldmath
  S3M: Supplementary Materials \\
  
\end{center}

\section{Correspondence Loss}
The functional loss consists of four components.
\begin{align}
    \label{eq:loss}
    \min_{\phi}\sum_{(X_i,X_j)} \mathcal{L}(C_{ij},C_{ji})&\text{\quad where \quad } C_{ij} = \argmin_{C}\lVert CA_{T_\phi(D_i)}-A_{T_\phi(D_j)} \rVert \\
    \mathcal{L} &= \mathcal{L}_{bij} + \mathcal{L}_{orth} + \mathcal{L}_{iso} + \mathcal{L}_{point} 
\end{align}

We first require the functional maps to be approximately bijective by regularizing their composition with the identity operator:
\begin{align}
     \mathcal{L}_{bij}(C_{ij},C_{ji}) = \lVert C_{ij}C_{ji} - I \rVert_F + \lVert C_{ji}C_{ij} - I \rVert_F
\end{align}

Furthermore, we enforce their orthogonality to maintain their local area preservation characteristics with $\mathcal{L}_{orth}$~\cite{ovsjanikovFunctionalMapsFlexible}, and a further regularization to ensure commutation with the LBO to maintain isometric properties on the surfaces with $\mathcal{L}_{iso}$:

\begin{align}
    \mathcal{L}_{orth}(C_{ij},C_{ji}) &= \lVert C_{ij}^T C_{ji} - I \rVert_F + \lVert C_{ji}^TC_{ij} - I \rVert_F\\
    \mathcal{L}_{iso}(C_{ij},C_{ji}) &= \lVert C_{ij} \Lambda_{i} - \Lambda_{j} C_{ij}  \rVert_F + \lVert C_{ji}\Lambda_j - \Lambda_i C_{ji} \rVert_F
\end{align}

\noindent where $\Lambda_{i}$ is the diagonal matrix of the LBO eigenvalues of shape $X_i$ and $\lVert \cdot \rVert_F$ denotes the frobenius norm.
Finally, in addition to regularization through these ``soft'' correspondences defined by the functional map, we further regularize ``hard'' point-to-point correspondences based on pointwise descriptors matched using the functional maps in the truncated LBO bases of each shape with $\mathcal{L}_{point}$ \cite{nogneng2017informative,roufosseUnsupervisedDeepLearning2019}.

\begin{figure}[htbp]
    \begin{adjustbox}{center}
    \includegraphics[width=0.80\textwidth]{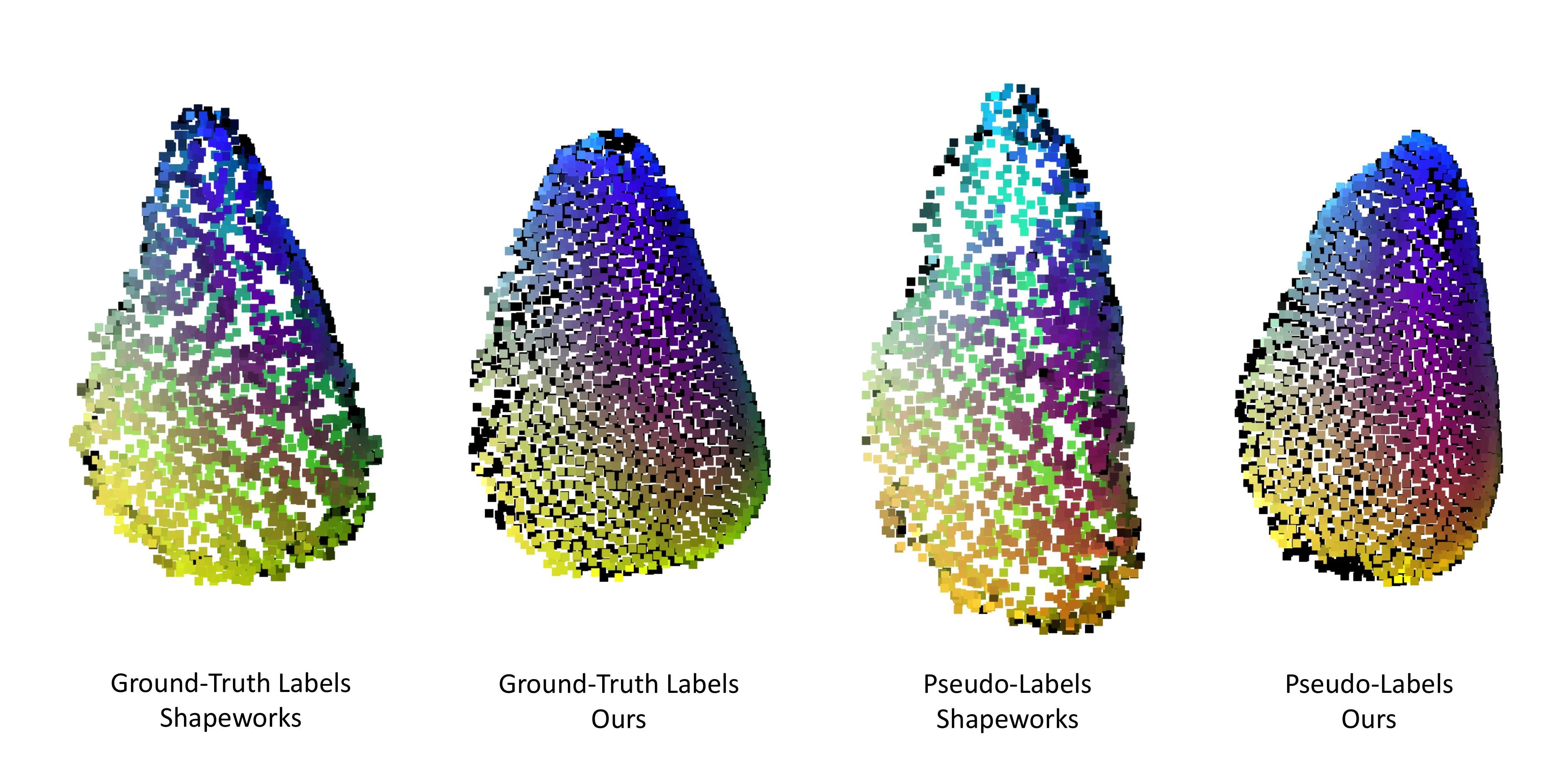}
    \end{adjustbox}
    \caption{Qualitative analysis of thyroid SSM mean shapes. Mean shapes on manually labeled shapes are depicted on the left, and on pseudo-labels from the QuickNat are on the right. }
    \label{fig:supp_thy}
\end{figure}

\begin{figure}[htbp]
    \begin{adjustbox}{center}
    \includegraphics[width=0.80\textwidth]{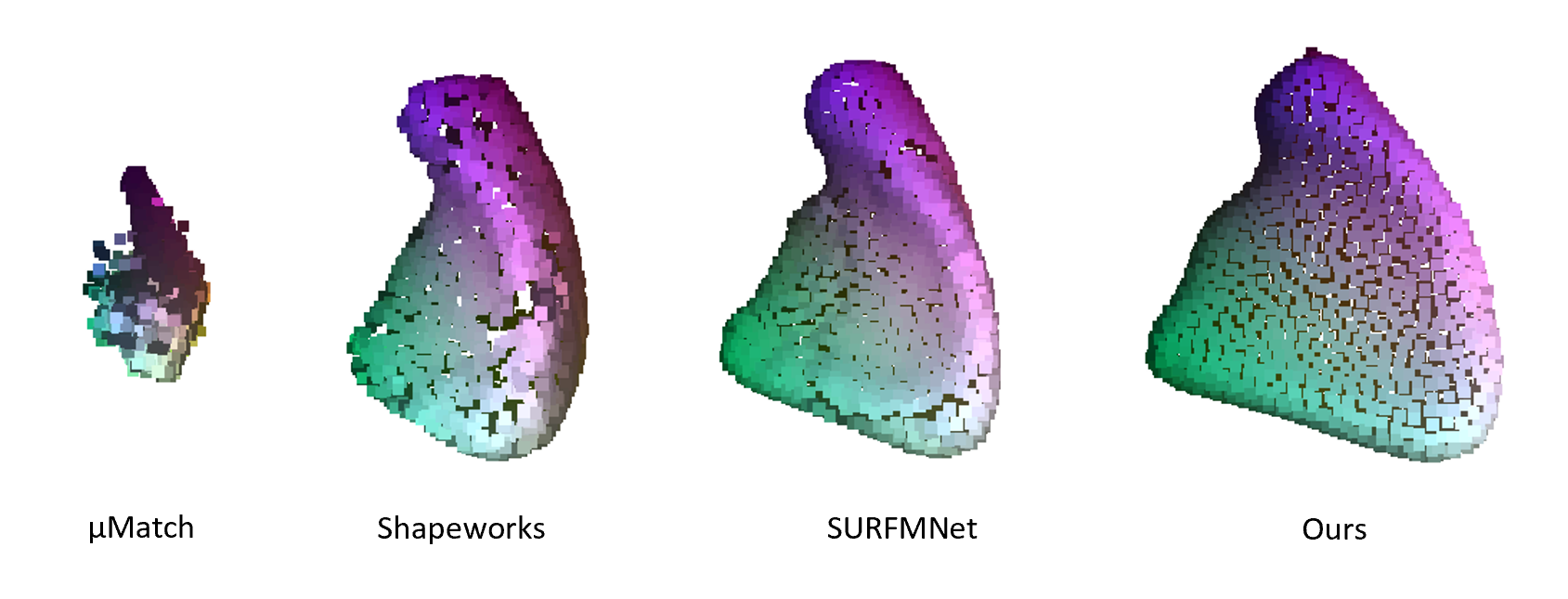}
    \end{adjustbox}
    \caption{Qualitative analysis of SSM mean shapes on Composition 1 (right ventricle) of heart dataset.}
    \label{fig:supp_h1}
\end{figure}

\begin{figure}[htbp]
    \begin{adjustbox}{center}
    \includegraphics[width=0.80\textwidth]{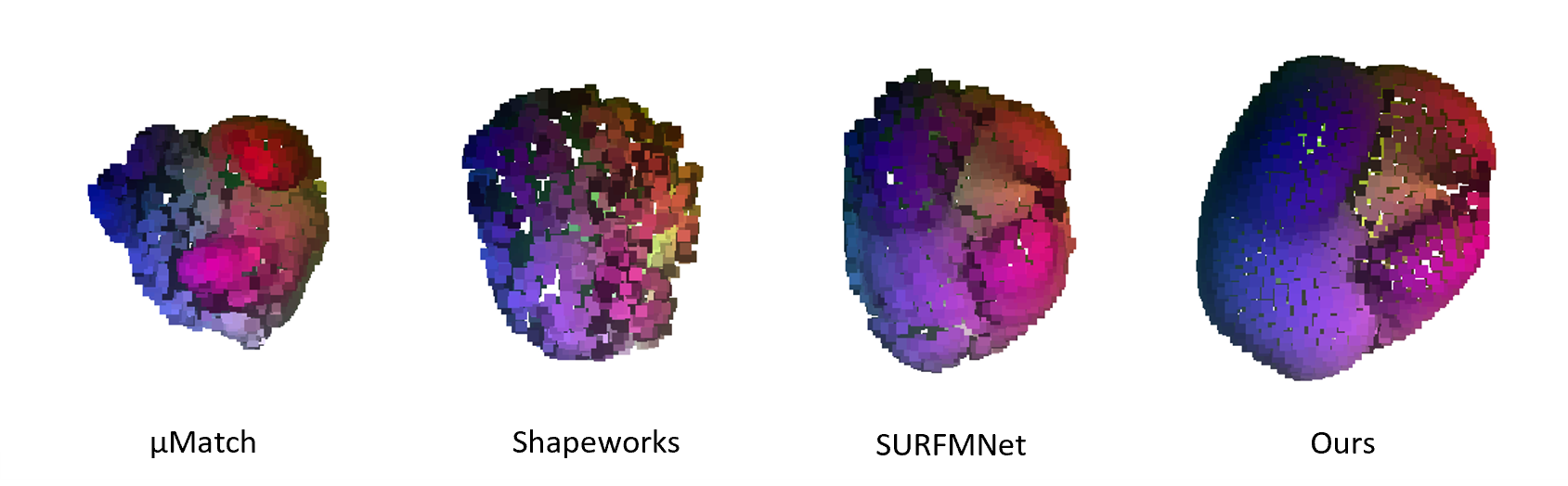}
    \end{adjustbox}
    \caption{Qualitative analysis of SSM mean shapes on Composition 2 of heart dataset.}
    \label{fig:supp_h3}
\end{figure}

\bibliographystyle{splncs04}
\bibliography{references}